\documentclass[nohyperref]{article}

\usepackage{microtype}
\usepackage{graphicx}
\usepackage{subfigure}
\usepackage{booktabs} %
\usepackage{multirow}

\usepackage{hyperref}

\newcommand{\V}[1]{{\mathbf{#1}}}

\usepackage[accepted]{icml2022}

\usepackage{amsmath}
\usepackage{amssymb}
\usepackage{mathtools}
\usepackage{amsthm}

\usepackage[capitalize,noabbrev]{cleveref}

\theoremstyle{plain}

\theoremstyle{definition}

\theoremstyle{remark}

\usepackage[textsize=tiny]{todonotes}

\usepackage{icml2022}
\usepackage{pifont}
\newcommand{\cmark}{\ding{51}}%
\newcommand{\xmark}{\ding{55}}%

\begin{document}

\twocolumn[
\icmltitle{LIMO: Latent Inceptionism for Targeted Molecule Generation}

\begin{icmlauthorlist}
\icmlauthor{Peter Eckmann}{cs-ucsd}
\icmlauthor{Kunyang Sun}{chemistry-ucsd}
\icmlauthor{Bo Zhao}{cs-ucsd}
\icmlauthor{Mudong Feng}{chemistry-ucsd}
\icmlauthor{Michael K. Gilson}{chemistry-ucsd,skaggs-ucsd}
\icmlauthor{Rose Yu}{cs-ucsd}
\end{icmlauthorlist}

\icmlaffiliation{cs-ucsd}{Department of Computer Science and Engineering, UC San Diego, La Jolla, California, United States}
\icmlaffiliation{chemistry-ucsd}{Department of Chemistry and Biochemistry, UC San Diego, La Jolla, California, United States}
\icmlaffiliation{skaggs-ucsd}{Skaggs School of Pharmacy and Pharmaceutical Sciences, UC San Diego, La Jolla, California, United States}

\icmlcorrespondingauthor{Michael Gilson}{mgilson@health.ucsd.edu}
\icmlcorrespondingauthor{Rose Yu}{roseyu@ucsd.edu}

\icmlkeywords{Machine Learning, ICML}

\vskip 0.3in
]

\printAffiliationsAndNotice{} %

\begin{abstract}
Generation of drug-like molecules with high binding affinity to target proteins remains a difficult and resource-intensive task in drug discovery. Existing approaches primarily employ reinforcement learning, Markov sampling, or deep generative models guided by Gaussian processes, which can be prohibitively slow when generating molecules with high binding affinity calculated by computationally-expensive physics-based methods. We present Latent Inceptionism on Molecules (LIMO), which significantly accelerates molecule generation with an inceptionism-like technique. LIMO employs a variational autoencoder-generated latent space and property prediction by two neural networks in sequence to enable faster gradient-based reverse-optimization of molecular properties. Comprehensive experiments show that LIMO performs competitively on benchmark tasks and markedly outperforms state-of-the-art techniques on the novel task of generating drug-like compounds with high binding affinity, reaching nanomolar range against two protein targets. We corroborate these docking-based results with more accurate molecular dynamics-based calculations of absolute binding free energy and show that one of our generated drug-like compounds has a predicted $K_D$ (a measure of binding affinity) of $6 \cdot 10^{-14}$ M against the human estrogen receptor, well beyond the affinities of typical early-stage drug candidates and most FDA-approved drugs to their respective targets. Code is available at \url{https://github.com/Rose-STL-Lab/LIMO}.
\end{abstract}
\section{Introduction}

Modern drug discovery is a long and expensive process, often requiring billions of dollars and years of effort \cite{hughes2011principles}. Accelerating the process and reducing its cost would have clear economic and human benefits. A central goal of the first stages of drug discovery, which comprise a significant fraction and cost of the entire drug discovery pipeline \cite{paul2010rdproductivity}, is to find a compound that has high binding affinity to a designated protein target, while retaining favorable pharmacologic and chemical properties \cite{hughes2011principles}. This task is difficult because there are on the order of $10^{33}$ chemically feasible molecules in the drug-like size range \cite{polishchuk2013estimation}, and only a tiny fraction of these bind to any given target with an affinity high enough to make them candidate drugs. Currently, this is done with large experimental compound screens and iterative synthesis and testing by medicinal chemists.

Recently, deep generative models have been proposed to identify promising drug candidates \cite{guimaraes2017objective, gomezbombarelli2018cvae, jin2018jtvae,ma2018constrained,you2018gcpn, popova2019mrnn,zhou2019moldqn,jin2020rationalerl,xie2021mars,luo2021graphdf}, potentially circumventing much of the customary experimental work. However, even the best generative methods are prohibitively slow when optimizing for molecular properties that are computationally expensive to evaluate, such as binding affinity. 

Here, we present a novel approach called Latent Inceptionism on Molecules (LIMO), a generative modeling framework for fast {\em de novo} molecule design that
\begin{itemize}
    \item builds on the variational autoencoder (VAE) framework, combined with a novel property predictor network architecture;
    \item employs an inceptionism-like reverse optimization technique on a latent space to generate drug-like molecules with desirable properties;
    \item is much faster than existing reinforcement learning-based methods ($6-8 \times$ faster) and sampling-based approaches ($12 \times$ faster), while maintaining or exceeding baseline performances on the generation of molecules with desired properties;
    \item allows for the generation of molecules with desired properties while keeping a molecular substructure fixed, an important task in lead optimization;
    \item markedly outperforms state-of-the-art methods in the novel task of generating drug-like molecules with high binding affinities to target proteins.
\end{itemize}
\section{Related Work}

\paragraph{Domain state of the art.}
After a protein is identified as a potential drug target, a common drug discovery paradigm today involves performing an initial high-throughput experimental screening of available compounds to identify hit compounds, i.e., molecules that have some affinity to the target. Computational methods, such as docking (e.g. \citet{santosmartins2021autodockgpu, friesner2004glide}) or more rigorous molecular dynamics-guided binding free energy calculations \cite{cournia_rigorous_2020} of compounds to a known 3D structure of the target protein can also play a role by prioritizing compounds for testing. Once hit compounds have been experimentally confirmed, they become starting points for the synthesis of chemically similar lead compounds that have improved activity but require further optimization (lead optimization) to become a drug candidate that is deemed promising enough to advance further through the drug discovery pipeline \cite{hughes2011principles}. To accelerate this often years-long drug discovery stage, there is great interest in novel computational technologies.

An alternative to experimentally screening existing compounds is to design entirely novel compounds for synthesis and testing. This approach, termed \textit{de novo} design, takes advantage of the target protein's 3D structure. Genetic algorithms (e.g. \citet{spiegel2020autogrow4}) and rule-based approaches (e.g. \citet{allen2017denovodock}) have been developed for this task. However, these techniques are often slow and tend to be too rigid to be fully integrated into the drug discovery process, where many molecular properties and synthesizability must be considered simultaneously. For example, AutoGrow4 \cite{spiegel2020autogrow4}, a state-of-the-art genetic algorithm, produces molecules with high binding affinities, but can also lead to toxic moieties and excessive molecular weights, while also being limited in the molecular space available for exploration. In contrast, recent machine learning methods offer greater flexibility and hence new promise for \textit{de novo} drug design \cite{carracedoreboredo2021mldrugdiscoveryreview}, as summarized below.

\paragraph{Generative models for molecule design.}
Deep generative models use a learned latent space to represent the distribution of drug-like molecules. Early work \cite{gomezbombarelli2018cvae} applies a variational autoencoder (VAE, \citet{kingma2013auto}) to map SMILES strings \cite{weininger1988smiles} to a continuous latent space. But SMILES-based representations struggle with generating both syntactically and semantically valid strings. Other works address this limitation by incorporating rules into the VAE decoder to only generate valid molecules \cite{kusner2017gvae, dai2018sdvae}. Junction tree VAEs \cite{jin2018jtvae, jin2019vtjnn} use a scaffold junction tree to assemble building blocks into an always-valid molecular graph, and have been improved with RL-like sampling and optimization techniques \cite{tripp2020sample-efficient, notin2021improving}. DruGAN \cite{kadurin2017drugan} further extends VAEs to an implicit GAN-based generative model. OptiMol uses a VAE to output molecular strings, but takes molecular graphs as input \cite{boitreaud2020optimol}. \citet{shen2021deepdreaming} forego a latent space altogether and assemble symbols directly.

Apart from sequence generation, graph generative models have also been proposed \cite{ma2018constrained, simonovsky2018graphvae, de2018molgan, li2018multi,fu2021dst, zang2020moflow, luo2021graphdf, jin2020hierarchical, luo20213d}. As generative models do not directly control molecular properties, existing methods often use a surrogate model (Gaussian process or neural network) to predict molecular properties from the latent space, and guide optimization on the latent space toward molecules with desired properties (e.g. logP, QED, binding affinity). For example, MoFlow \cite{zang2020moflow} predicts molecular properties from a latent space using a neural network, but has difficulty generating molecules with high property scores. Instead, we propose the prediction of properties from the {\em decoded} molecular space, which appears to greatly increase the property scores of generated molecules. \citet{xie2021mars} propose Monte Carlo sampling to explore molecular space and \citet{nigam2020gad} propose a genetic algorithm with a neural network-based discriminator, both of which require an extremely large number of calls to property functions and therefore are less useful when optimizing complex, expensive-to-evaluate property functions.

In general, generative models are very fast in generating molecules. However, as current generative models cannot effectively find molecules in their latent spaces that have desired properties, they have so far been outperformed by reinforcement learning-based methods that directly optimize molecules for desired properties.

\paragraph{Reinforcement learning-based molecule generation.}
Reinforcement learning (RL) methods directly optimize molecular properties by systematically constructing or altering a molecular graph \cite{you2018gcpn, zhou2019moldqn, jin2020rationalerl, guimaraes2017objective, popova2019mrnn, de2018molgan, zhavoronkov2019gentrl, olivecrona2017molecular, shi2020graphaf, luo2021graphdf, jeon2020autonomous}. These methods appear to be the most powerful at generating molecules with desired properties, but are slow and require many calls to the property estimation function. This is problematic when applying RL to computationally expensive but highly useful property functions like physics-based (e.g. docking) computed binding affinity, rather than simple, easily computed measures such as logP. RationaleRL \cite{jin2020rationalerl} theoretically avoids the need to sample a large number of molecules by collecting ``rationales'' from existing molecules with desired properties, and combining them into molecules with multiple desired properties, but by design this method is not applicable to {\em de novo} drug discovery.
\section{Methodology}

\begin{figure*}[t]
\begin{center}
\centerline{\includegraphics[width=\textwidth]{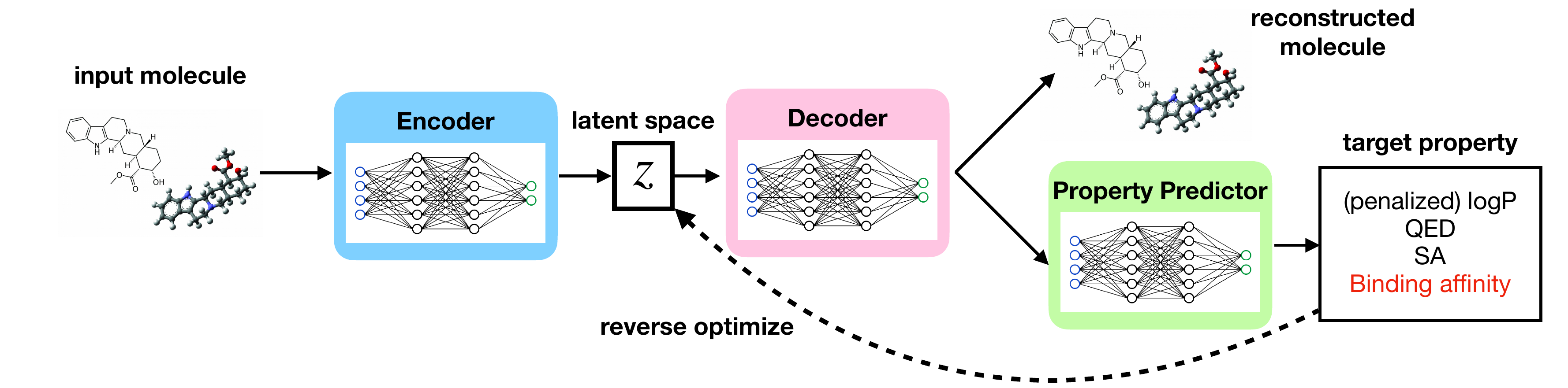}}
\caption{Overview of the LIMO framework. We train a variational autoencoder (``Encoder'' and ``Decoder'') to reconstruct  input drug-like molecules. Then, a property predictor is trained to predict molecular properties (``target property'') using the output of the decoder. Using the property predictor, we generate molecules with desired properties by performing gradient descent on the output of the property predictor with respect to the latent space $\V{z}$, an inceptionism-like approach.}
\label{pipeline}
\end{center}
\vskip -0.1in
\end{figure*}

We present Latent Inceptionism on Molecules (LIMO), a molecular generation framework. We use a VAE to learn a real-valued latent space representation of the drug-like chemical space. However, contrary to previous work, we use two neural networks (a decoder and predictor) in sequence to perform inceptionism-like reverse optimization of molecular properties on the latent space. Figure \ref{pipeline} gives an overview of the framework.

We use a decoder network to generate an intermediate real-valued molecular representation to improve the prediction, and therefore optimization, of molecular properties while keeping the prediction differentiable, allowing the use of efficient gradient-based optimizers. We use self-referencing embedded strings (SELFIES, \citet{krenn2020selfies}) to ensure chemical validity during optimization. With these novelties, LIMO is able to achieve performance on par with reinforcement learning methods while being orders of magnitude faster. On the highly useful task of structure-based computed binding affinity optimization, LIMO markedly outperforms state-of-the-art (including RL) methods, while also being much faster.

\subsection{Variational Autoencoder} \label{methods-autoencoder}

Define a string representation of a molecule $\V{x} = (x_1, \cdots, x_n)$. Each $x_i$ takes its value from a set of $d$ possible symbols $\mathcal{S} = \{s_1, \cdots, s_d\}$, where each symbol is one component of a self-referencing embedded string (SELFIES) defining a molecule \cite{krenn2020selfies}. We aim to produce $n$ independent distributions $\V{y} = (y_1, \cdots, y_n)$, where each $y_i \in [0, 1]^d$ is the parameter for a multinomial distribution over the $d$ symbols in $\mathcal{S}$. The output string $\V{\hat{x}}= (\hat{x}_1, \cdots, \hat{x}_n)$ is obtained from $\V{y}$ by selecting the symbol with the highest probability:
\begin{align} \label{eq:1}
    \hat{x}_i = s_{d^*_i}, \quad d^*_i=\text{argmax}_{d} \{y_{i,1},\cdots, y_{i,d}\}
\end{align}
All SELFIES strings correspond to valid molecules, allowing us to transform the continuous-value probabilities $\V{y}$ into an always-valid discrete molecule.

We train a VAE \cite{kingma2013auto} to encode  $\V{x}$ to a latent space $\V{z} \in \mathbb{R}^{m}$ of dimensionality $m$ and decode to $\V{y}$. We optimize the VAE using the evidence lower bound (ELBO) loss function. Each input symbol in the representation string passes through an embedding layer, and then two fully-connected networks (the encoder and decoder). Reconstruction loss is calculated using the negative log-likelihood over a one-hot encoded representation of the input molecule. Once trained, the VAE can generate novel and drug-like molecules, with similar molecules lying next to each other in the latent space. To generate random molecules, we sample from the latent space $\V{z}$ using $\mathcal{N}(0_m, I_m)$, and decode it into a string representation.

\subsection{Property Predictor} \label{methods-prop-prediction}

We employ a separate network to predict molecular properties. While earlier works train the VAE and property predictor jointly \cite{jin2018jtvae, gomezbombarelli2018cvae}, we train the property predictor {\em after} the VAE has been fully trained (i.e. we freeze the VAE weights) for three reasons: firstly, generative modeling requires significantly more molecular data than the regression task of predicting molecular properties.There is no need to acquire property data for all the molecules used by the generative model. This is especially relevant when such data is expensive to obtain, e.g. docking-based binding affinity that takes seconds to calculate per molecule. Secondly, the trained generative model allows us to query the ground-truth molecular property function with its generated molecules, giving an informative and diverse training set for property prediction. Thirdly, adding new properties under this training scheme does not require retraining of the VAE, only of the property predictor, which is much more efficient.

Crucially, we introduce a novel architecture consisting of stacking the VAE decoder and the property predictor. The property predictor uses the output of the VAE decoder as its input, as opposed to predicting properties directly from the latent space like previous works (e.g. \citet{jin2018jtvae, gomezbombarelli2018cvae, zang2020moflow}). The intuition is that the map from molecular space to property is easier to learn than that from the latent space to property. We later present results confirming this intuition, both in terms of prediction accuracy and overall optimization ability, suggesting that the proposed stacking improves optimization by allowing more accurate prediction of molecular properties through a more direct molecular representation. Using such an intermediate molecular representation from the VAE decoder also allows us to fix a substructure of the generated molecule, giving LIMO the ability to perform the unique, compared to many other VAE-based architectures, ability to perform substructure-constrained optimization.

Define the VAE encoder $f_\text{enc}: \V{x} \mapsto \V{z}$ and decoder $f_\text{dec}: \V{z} \mapsto \V{\hat{x}}$, a property prediction network $g_\theta: \V{\hat{x}} \mapsto \mathbb{R}$ with parameters $\theta$, and a ground-truth property estimation function $\pi: \V{\hat{x}} \mapsto \mathbb{R}$ that computes a molecular property such as logP or binding affinity. We first generate examples to train $g_\theta$ by sampling random molecules from the latent space $\V{z}$ using a normal distribution $\mathcal{N}(0_m, I_m)$. Then, we optimize the parameters $\theta$ of the property predictor by minimizing the mean square error (MSE) of predicted properties over the set of generated molecules:
\vspace{-0.2cm}
\begin{align}
\ell_0 (\theta) = \|g_\theta(f_\text{dec}(\V{z})) - \pi(f_\text{dec}(\V{z}))\|^2
\end{align}

\subsection{Reverse Optimization}
\label{sec:reverse-optimization-methods}
After training, we freeze the weights of $f_\text{dec}$ and $g_\theta$, and make $\V{z}$ trainable to optimize the latent space toward locations that decode to molecules with desirable properties. This is a similar technique to inceptionism, which involves backpropagating from the output of a network to its input so that the input is altered to affect the output in a desired way \cite{mordvintsev2015inceptionism}. 

To maximize properties, given a set of $k$ property predictors $(g^1, \cdots, g^k)$ and weights $(w_1, \cdots, w_k)$, we minimize the following function using the Adam optimizer, initialized from $\V{z} \sim \mathcal{N}(0_m, I_m)$:
\begin{align} \label{eq:2}
\ell_1(\V{z}) = -\sum_{i=1}^{k} w_i \cdot g^i(f_\text{dec}(\V{z}))
\end{align}
Crucially, since both $f_\text{dec}$ and $g$ are neural networks, gradient-based techniques can be used for efficient optimization of $\V{z}$. The weights $(w_1, \cdots, w_k)$ are hyperparameters determined by a random grid search.

In lead optimization, a common task is to generate molecules with desired properties while keeping a given substructure of the molecules fixed. To apply reverse optimization to this task, we define a mask $M \in \{0, 1\}^{n \times d}$, where $M_{i, j}$ corresponds to the SELFIES symbol of index $j$ at position $i$ in a molecular string. We assign $M_{i, j} = 1$ where the desired substructure is present and the corresponding symbol cannot be changed, 0 otherwise. For an optimization starting point, we then reconstruct a molecule $\V{x}_\text{start}$ that has the desired substructure: $\V{\hat{x}}_\text{start} = f_\text{dec}(f_\text{enc}(\V{x}_\text{start}))$. To optimize $\V{z}$ while also keeping a substructure constant, we add an additional loss $\ell_2$ to the $\ell_1$ used in Equation \ref{eq:2}:
\vspace{-0.3cm}
\begin{align} \label{eq:4}
    \ell_2(\V{z}) = \lambda \sum_{i=1}^{n} \sum_{j=1}^{d} (M_{i,j} \cdot (f_\text{dec}(\V{z})_{i,j} - (\V{\hat{\V{x}}}_\text{start})_{i,j}))^2
\end{align}
where $\lambda$ is a weighting term we set to 1,000.

\subsection{Refinement} \label{sec:refinement}
\paragraph{Filtering.}
Following multi-objective optimization, we perform a filtering step to exclude non drug-like molecules. Using the distributions of quantitative estimate of drug-likeness (QED) and synthetic accessibility (SA) scores on drug-like datasets \cite{bickerton2012qed, ertl2009sa}, we define cutoff values reasonably within the range of currently marketed drugs. Molecules not reaching these cutoffs are excluded from consideration. We also exclude molecules with either too small or too large chemical cycles (rings), as these are usually difficult to synthesize but are not excluded effectively by the SA metric. Specifically, we exclude molecules not satisfying $(\text{QED}>0.4) \wedge (\text{SA}<5.5) \wedge (\text{no rings with } <5 \text{ or } >6 \text{ atoms})$.

\paragraph{Fine-tuning.}
For some tasks, we observe that LIMO is effective in generating molecules with reasonably high property scores that could nonetheless be improved slightly by small, atom-level changes. To do this, we performed a greedy local search around the chemical space of a generated molecule by systematically replacing carbons with heteroatoms and retaining changes that lead to the most improvement. The algorithm is detailed in Appendix \ref{sec:appendix-finetuning}.
\section{Experiments}

We apply LIMO to QED and penalized logP (p-logP) maximization, logP targeting \cite{you2018gcpn}, similarity-constrained p-logP maximization \cite{jin2018jtvae}, substructure-constrained logP extremization, and single and multi-objective binding affinity maximization. All of these tasks are typical challenges in drug discovery, especially optimization around a substructure and maximization of binding affinity. See Appendix \ref{sec:appendix-tasks} for description of each task, and Appendix \ref{sec:appendix-random-generation} for results from the random generation of molecules.

\subsection{Experimental Setup} \label{sec:experimental-setup}
\paragraph{Dataset.}
For all optimization tasks, we use the benchmark ZINC250k dataset, which contains $\approx$250,000 purchasable, drug-like molecules \cite{irwin2012zinc}. We use AutoDock-GPU \cite{santosmartins2021autodockgpu} to compute binding affinity, as described in Appendix \ref{sec:autodock-methods}, and RDKit to compute other molecular properties. For the random generation task, we train on the ZINC-based $\approx$2 million molecule MOSES dataset \cite{polykovskiy2020moses}.

\paragraph{Model training.}
All experiments use identical autoencoder weights and a latent space dimension of 1024. We select hyperparameters using a random grid search. The property predictor is trained independently for each of the following properties: logP (octanol-water partition coefficient), p-logP \cite{jin2018jtvae}, SA \cite{ertl2009sa}, QED \cite{bickerton2012qed}, and binding affinity to two targets (calculated by AutoDock-GPU, \citet{santosmartins2021autodockgpu}). 100k training examples were used for all properties except binding affinity, where 10k were used due to speed concerns. See Appendix \ref{sec:experimental-details} for model training details.

\paragraph{Baselines.}
We compare with the following state-of-the-art molecular design baselines: JT-VAE \cite{jin2018jtvae}, GCPN \cite{you2018gcpn}, MolDQN \cite{zhou2019moldqn}, MARS \cite{xie2021mars}, and GraphDF \cite{luo2021graphdf}. Each technique is described in Appendix \ref{sec:appendix-baselines}.

\paragraph{Protein targets.}

For tasks involving binding affinity optimization, we target the binding sites of two human proteins:

\begin{itemize}
    \item \textbf{Human estrogen receptor (ESR1)}: This well-characterized protein is a target of drugs used to treat breast cancer. It was chosen for its disease relevance and its many known binders, which are good points of comparison with generated molecules. Although known binders exist, LIMO was not fed any information beyond a crystal structure of the protein (PDB 1ERR) used for docking calculations and the location of the binding site.
    \item \textbf{Human peroxisomal acetyl-CoA acyl transferase 1 (ACAA1)}: This enzyme has no known binders but does have a crystal structure (PDB 2IIK) with a potential drug-binding pocket, which we target to show the ability of LIMO for {\em de novo} drug design. We found this protein with the help of the Structural Genomics Consortium, which highlighted this protein as a potentially disease-relevant target with a known crystal structure, but no known binders.
\end{itemize}

\subsection{QED and Penalized logP Maximization}

\begin{table*}[ht]
    \vskip -0.1in
    \caption{Comparison of QED and p-logP maximization methods. ``LL'' (length limit) denotes whether a model has a limited output length (about the maximum molecule size of ZINC250k), as p-logP score can increase linearly with molecule length. Baseline results taken from \cite{you2018gcpn, luo2021graphdf, xie2021mars}.}
    \label{property-optimization}
    \vskip 0.1in
    \begin{center}
    \begin{small}
    \begin{sc}
    \begin{tabular}{l|c|ccc|ccc|c}
    \toprule
    Method & LL & \multicolumn{3}{c}{penalized logP} & \multicolumn{3}{c}{QED} & Time\\
    & & 1st & 2nd & 3rd & 1st & 2nd & 3rd & (hrs)\\
    \midrule
    JT-VAE & \xmark & 5.30 & 4.93 & 4.49 & 0.925 & 0.911 & 0.910 & 24\\
    GCPN & \cmark & 7.98 & 7.85 & 7.80 & \textbf{0.948} & 0.947 & 0.946 & 8\\
    MolDQN & \cmark & 11.8 & 11.8 & 11.8 & \textbf{0.948} & 0.943 & 0.943 & 24\\
    MARS & \xmark & \textbf{45.0} & \textbf{44.3} & \textbf{43.8} & \textbf{0.948} & \textbf{0.948} & \textbf{0.948} & 12\\
    GraphDF & \xmark & 13.7 & 13.2 & 13.2 & \textbf{0.948} & \textbf{0.948} & \textbf{0.948} & 8\\
    \midrule
    LIMO on $\V{z}$ & \cmark & 6.52 & 6.38 & 5.59 & 0.910 & 0.909 & 0.892 & \textbf{1}\\
    LIMO & \cmark & 10.5 & 9.69 & 9.60 & 0.947 & 0.946 & 0.945 & \textbf{1}\\
    \bottomrule
    \end{tabular}
    \end{sc}
    \end{small}
    \end{center}
    \vskip -0.1in
    \end{table*}

Table \ref{property-optimization} shows results of LIMO and baselines on the generation of molecules with high penalized logP and QED scores. For both properties, we report the top 3 scores of 100k generated molecules, as well as the total time (generation + testing) taken by each method. As an ablative study, we apply LIMO with property prediction directly on the latent space (``LIMO on $\V{z}$'') as opposed to regular LIMO, which performs property prediction on the decoded molecule $\V{\hat{x}}$ (see Section \ref{methods-prop-prediction}). Both methods underwent the same hyperparameter tuning as described in Appendix \ref{sec:experimental-details}. We see that the extra novel step of decoding the latent space and then performing property prediction offers a significant advantage for the optimization of molecules. To elucidate this improvement, an unseen test set of 1,000 molecules was generated using the VAE and used to test the prediction performance of the property predictor. We observe an $r^2 = 0.04$ between real and predicted properties for ``LIMO on $z$'', and $r^2 = 0.38$ for LIMO. This large predictive performance boost explains the observed improvements in the optimization of molecules, as the model is better able to generalize what makes a molecule bind well. We also replaced LIMO's fully-connected VAE encoder and decoder each with an 8-layer, 512 hidden dimension LSTM and found significantly worse performance, e.g. a maximum QED score of 0.3. The addition of a self-attention layer after the LSTM encoder did not significantly improve performance.

We observe that LIMO achieves competitive results among deep generative and RL-based models (i.e. all methods except MARS) while taking significantly less time. Note that p-logP is a ``broken'' metric that is almost entirely dependent on molecule length \cite{zhou2019moldqn}. Without a length limit, MARS can easily generate long carbon chains with high p-logP. Among models with a molecule length limit (GCPN, MolDQN, and LIMO), LIMO generates molecules with p-logP similar to MolDQN, the strongest baseline. Similarly, QED suffers from boundary effects around its maximum score of 0.948 \cite{zhou2019moldqn}, which LIMO gets very close to. Drugs with a QED score above 0.9 are very rare \cite{bickerton2012qed}, so achieving close to this maximum score is sufficient for drug discovery purposes.

\subsection{logP Targeting}

Table \ref{property-targeting} reports on the ability of LIMO to generate molecules with logP targeted to the range $-2.5 < \text{logP} < -2.0$. LIMO achieves the highest diversity among generated molecules within the targeted logP range, and, although it has a lower success rate than other methods, it generates 33 molecules per second within the target range. This is similar to the overall generation speed of other models, but due to a lack of available code for this task, we were not able to compare exact speeds.
\vskip -2mm
\begin{table}[ht]
    \caption{Property targeting to $-2.5 < \text{logP} < -2.0$. Success (\%): percent of generated molecules within the target range. Diversity: One minus the average pairwise Tanimoto similarity between Morgan fingerprints. Results for JT-VAE and GCPN taken from \cite{you2018gcpn}.}
    \label{property-targeting}
    \vskip 0.1in
    \begin{center}
    \begin{small}
    \begin{sc}
    \begin{tabular}{lccccccr}
    \toprule
    Method & Success (\%) & Diversity\\
    \midrule
    JT-VAE & 11.3 & 0.846\\
    GCPN & \textbf{85.5} & 0.392\\
    MolDQN & 9.66 & 0.854\\
    GraphDF & 0 & -\\
    \midrule
    LIMO & 10.4 & \textbf{0.914}\\
    \bottomrule
    \end{tabular}
    \end{sc}
    \end{small}
    \end{center}
    \vskip -0.1in
    \end{table}
    
\begin{table*}[t!]
    \vskip -0.1in
    \caption{Similarity-constrained p-logP maximization. For each method and minimum similarity constraint $\delta$, the mean $\pm$ standard deviation of improvement (among molecules that satisfy the similarity constraint) from the starting molecule is shown, as well as the percent of optimized molecules that satisfy the similarity constraint (\% succ.). Baseline results taken from \cite{luo2021graphdf, zhou2019moldqn}.}
    \label{constrained-optimization}
    \vskip 0.1in
    \tabcolsep=0.15cm
    \begin{center}
    \begin{small}
    \begin{sc}
    \begin{tabular}{c|cc|cc|cc|cc|cc}
    \toprule
    \multirow{2}{1em}{$\delta$} & \multicolumn{2}{c}{JT-VAE} & \multicolumn{2}{c}{GCPN} & \multicolumn{2}{c}{GraphDF} & \multicolumn{2}{c}{MolDQN} & \multicolumn{2}{c}{LIMO}\\
    & Improv. & \% Succ. & Improv. & \% Succ. & Improv. & \% Succ. & Improv. & \% Succ. & Improv. & \% Succ.\\
    \midrule
    0.0 & $1.9 \pm 2.0$ & 97.5 & $4.2 \pm 1.3$ & \textbf{100} & $5.9 \pm 2.0$ & \textbf{100} & $7.0 \pm 1.4$ & \textbf{100} & $\boldsymbol{10.1 \pm 2.3}$ & \textbf{100}\\
    0.2 & $1.7 \pm 1.9$ & 97.1 & $4.1 \pm 1.2$ & \textbf{100} & $5.6 \pm 1.7$ & \textbf{100} & $5.1 \pm 1.8$ & \textbf{100} & $\boldsymbol{5.8 \pm 2.6}$ & 99.0\\
    0.4 & $0.8 \pm 1.5$ & 83.6 & $2.5 \pm 1.3$ & \textbf{100} & $\boldsymbol{4.1 \pm 1.4}$ & \textbf{100} & $3.4 \pm 1.6$ & \textbf{100} & $3.6 \pm 2.3$ & 93.7\\
    0.6 & $0.2 \pm 0.7$ & 46.4 & $0.8 \pm 0.6$ & \textbf{100} & $1.7 \pm 1.2$ & 93.0 & $\boldsymbol{1.9 \pm 1.2}$ & \textbf{100} & $1.8 \pm 2.0$ & 85.5\\
    \bottomrule
    \end{tabular}
    \end{sc}
    \end{small}
    \end{center}
    \vskip -0.1in
    \end{table*}

\subsection{Similarity-constrained Penalized logP Maximization}

Following the procedures described for JT-VAE \cite{jin2018jtvae}, we select the 800 molecules with the lowest p-logP scores in the ZINC250k dataset and aim to generate new molecules with a higher p-logP yet similarity to the original molecule. Similarity is measured by Tanimoto similarity between Morgan fingerprints with a cutoff value $\delta$. Each of the 800 starting molecules are encoded into the latent space using the VAE encoder, 1,000 gradient ascent steps (Section \ref{sec:reverse-optimization-methods}) are completed for each, then the generated molecules out of all gradient ascent steps with the highest p-logP that satisfy the similarity constraint are chosen.

Results for the similarity-constrained p-logP maximization task are summarized in Table \ref{constrained-optimization}. For the two lowest similarity constraints ($\delta = 0.0, 0.2$), LIMO achieves the highest penalized logP improvement, while its improvement is statistically indistinguishable from other methods at higher values of $\delta$. This shows the power of LIMO for unconstrained optimization, and the ability to reach competitive performance in more constrained settings.

\subsection{Substructure-constrained logP Extremization}

\begin{figure}[h]
\vskip -0.1in
\begin{center}
\centerline{\includegraphics[width=\columnwidth]{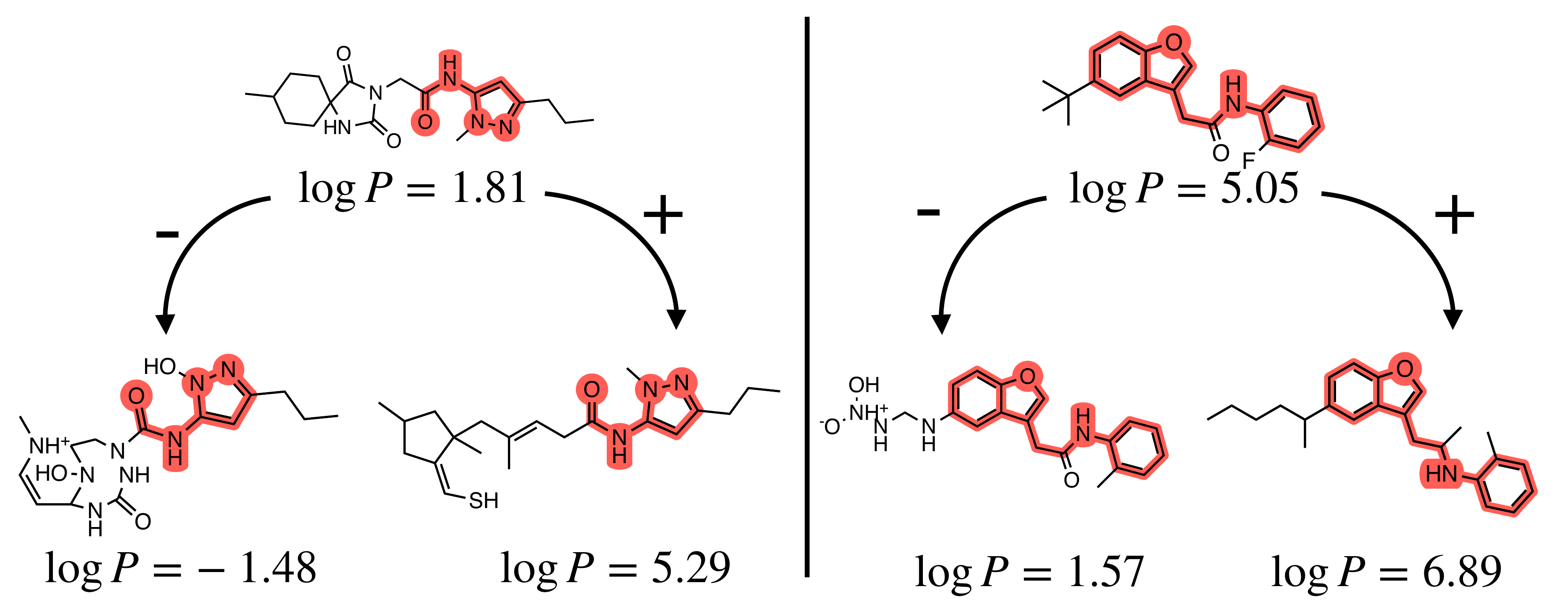}}
\caption{Examples of LIMO's extremization of logP while keeping a substructure (denoted in red) constant.}
\label{substructure-optimization}
\end{center}
\vskip -0.2in
\end{figure}

Results for the substructure-constrained logP extremization task are shown in Figure \ref{substructure-optimization}. We chose two molecules from ZINC250k to act as starting molecules and defined the substructures of these starting molecules to be fixed, then performed both maximization and minimization of logP using LIMO, as described in Equation \ref{eq:4}. As illustrated, we can successfully increase or decrease logP as desired while keeping the substructure constant in both cases.

This task is common during the lead optimization stage of drug development, where a synthetic pathway to reach an exact substructure with proven activity is established, but molecular groups around this substructure are more malleable and have not yet been determined. This is not captured in the similarity-constrained optimization task above, which uses more general whole-molecule similarity metrics.

While previous works address the challenge of property optimization around a fixed substructure \cite{hataya2021graph, lim2020scaffold, maziarz2022learning}, LIMO is one of the few VAE-based methods that can easily perform such optimization. Thanks to its unique decoding step of generating an intermediate molecular string prior to property prediction, LIMO brings the speed benefits of VAE techniques to the substructure optimization task.

\subsection{Single-objective Binding Affinity Maximization}
\label{sec:single-obj}
\begin{table}[h]
\vskip -0.11in
\caption{Generation of molecules with high computed binding affinities (shown as dissociation constants, $K_D$, in nanomoles/liter) for two protein targets, ESR1 and ACAA1.}
\label{single-obj-binding}
\vskip 0.1in
\tabcolsep=0.15cm
\begin{center}
\begin{small}
\begin{sc}
\begin{tabular}{l|ccc|ccc|c}
\toprule
\multirow{2}{2em}{Method} & \multicolumn{3}{c}{ESR1} & \multicolumn{3}{c}{ACAA1} & \multirow{2}{2em}{Time (hrs)}\\
& 1st & 2nd & 3rd & 1st & 2nd & 3rd &\\
\midrule
GCPN & 6.4 & 6.6 & 8.5 & 75 & 83 & 84 & 6\\
MolDQN & 373 & 588 & 1062 & 240 & 337 & 608 & 6\\
GraphDF & 25 & 47 & 51 & 370 & 520 & 590 & 12\\
MARS & 17 & 64 & 69 & 163 & 203 & 236 & 6\\
\midrule
LIMO & \textbf{0.72} & \textbf{0.89} & \textbf{1.4} & \textbf{37} & \textbf{37} & \textbf{41} & \textbf{1}\\
\bottomrule
\end{tabular}
\end{sc}
\end{small}
\end{center}
\vskip -0.15in
\end{table}

\begin{figure}[h]
\begin{center}
\centerline{\includegraphics[width=\columnwidth]{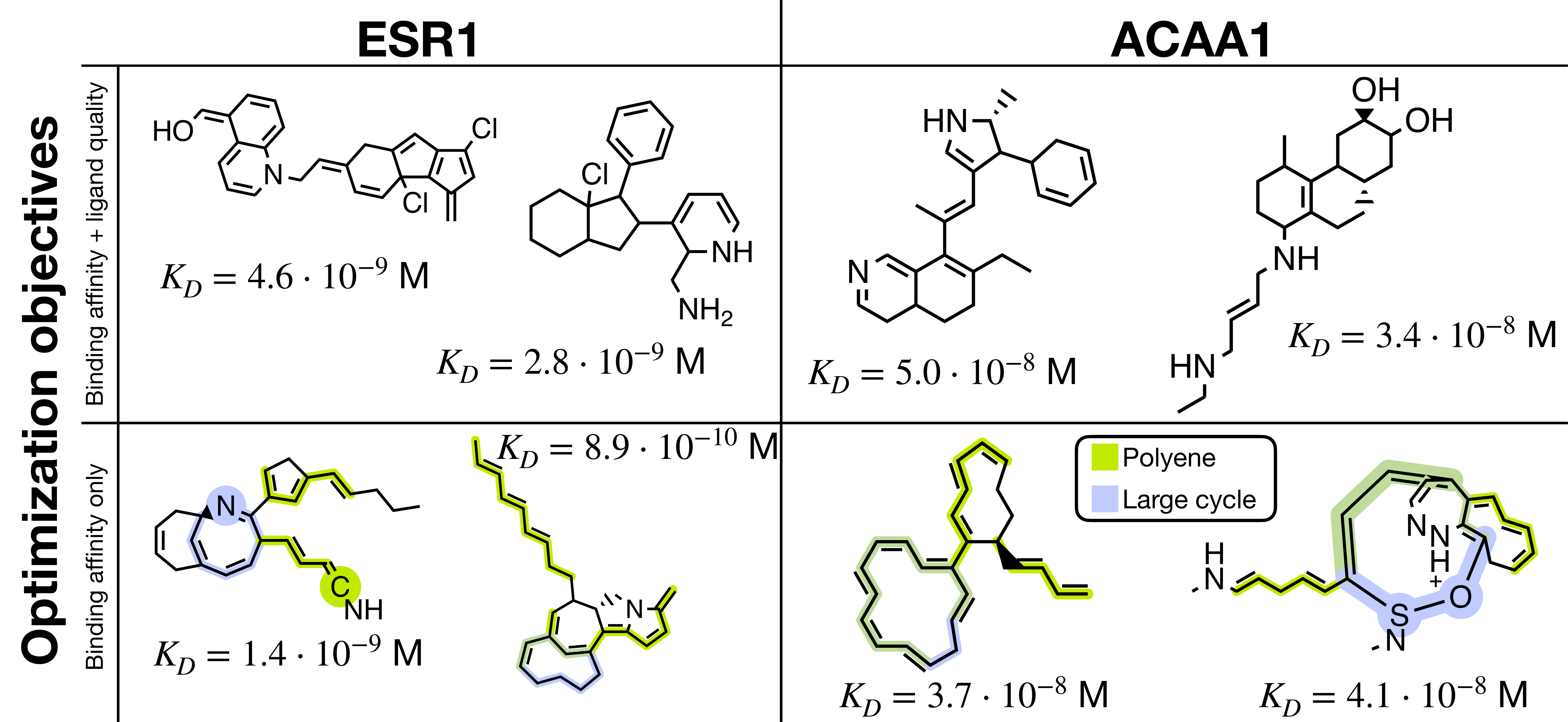}}
\caption{Generated molecules from the multi-objective (top row) and single-objective (bottom row) binding affinity maximization. The estimated dissociation constants, $K_D$, were obtained by docking each compound to the targeted protein using AutoDock-GPU. The dissociation constant is a measure of binding affinity, where lower is better. In the bottom row, we highlight two major problematic patterns that appeared when only considering computed binding affinity, motivating multi-objective optimization.}
\label{mol-examples}
\end{center}
\vskip -0.2in
\end{figure}

Producing molecules with high binding affinity to the target protein is the primary goal of early drug discovery \cite{hughes2011principles}, and its optimization using a docking-based binding affinity estimator, which is especially powerful in the {\em de novo} setting, is relatively novel to the ML-based molecule generation literature. Many previous approaches have attempted to optimize affinity by leveraging knowledge of existing binders (e.g. \citet{zhavoronkov2019gentrl, jeon2020autonomous, luo20213d}), but they often lack generalizability to targets without such binders. Therefore, we focus on molecule optimization in the {\em de novo} setting through the use of a docking-based affinity estimator.%

We target the binding sites of two human proteins: estrogen receptor (PDB ESR1, UniProt P03372) and peroxisomal acetyl-CoA acyl transferase 1 (PDB ACAA1, UniProt P09110) (see Section \ref{sec:experimental-setup} for details). For both of our protein targets we report the top 3 highest affinities (i.e., lowest dissociation constants, $K_D$, as estimated with AutoDockGPU) of 10k total generated molecules from each method. As shown in Table \ref{single-obj-binding}, LIMO generates compounds with higher computed binding affinities in far less time than prior state-of-the-art methods. We chose GCPN, MolDQN, GraphDF, and MARS as baseline comparisons because of their strong performance on other single-objective optimization tasks.

The chemical structures of two molecules generated by LIMO when only optimizing for binding affinity are shown in the bottom row of Figure \ref{mol-examples} for both protein targets. While these molecules have relatively high affinities, they would have little utility in drug discovery because they are pharmacologically and synthetically problematic. For example, we highlight two major moieties, polyenes and large ($\geq$8 atoms) cycles, that are regarded by domain experts as highly problematic due to reactivity/toxicity and synthesizability concerns, respectively (see \citet{birch2017polyenes, hussain2014macrocycles, abdelraheem2016macrocycles2}). Molecules generated from GCPN, MolDQN, GraphDF, and MARS had similar issues. These moieties are large structural issues that cannot be fixed with small tweaks following optimization, so we added measures of ligand quality into our optimization process as detailed in the following subsection.

\subsection{Multi-objective Binding Affinity Maximization}

\begin{table*}[h]
\vskip -0.1in
\caption{Comparison of generated ligands for ESR1 and ACAA1 following multi-objective optimization and refinement. Arrows indicate whether a high score ($\uparrow$) or low score ($\downarrow$) is desired. High QED, $\text{Fsp}^3$, and satisfying Lipinski's Rule of 5 suggest drug-likeness. A low number of PAINS alerts indicates a low likelihood of false positive results in binding assays. MCE-18 is a measure of molecular novelty based on complexity, and SA is a measure of synthesizability. $K_D$ values in nM are computed binding affinities from AutoDock-GPU (AD) and from more rigorous absolute binding free energy calculations (ABFE). See Appendix \ref{sec:molecule-metrics} for a full description of each metric. * indicates an experimentally determined value obtained from BindingDB \cite{liu2007bindingdb}.}
\label{multi-obj-comparison}
\vskip 0.1in
\tabcolsep=0.1cm
\begin{center}
\begin{small}
\begin{sc}
\begin{tabular}{l|ccc|ccccc}
\toprule
\multirow{2}{2em}{Ligand} & \multicolumn{3}{c}{Optimized prop.} & \multicolumn{5}{c}{Non-optimized prop.}\\
& $K_D$ (AD) ($\downarrow$) & QED ($\uparrow$) & SA ($\downarrow$) & $K_D$ (ABFE) ($\downarrow$) & Lipinski & PAINS ($\downarrow$) & $\text{Fsp}^3$ ($\uparrow$) & MCE-18 ($\uparrow$)\\
\midrule
\multicolumn{9}{c}{ESR1}\\
\midrule
LIMO mol. \#1 & 4.6 & 0.43 & 4.8 & $\boldsymbol{6 \cdot 10^{-5}}$ & \cmark & 0 & 0.16 & 90\\
LIMO mol. \#2 & \textbf{2.8} & 0.64 & 4.9 & 1000 & \cmark & 0 & 0.52 & 76\\
\midrule
GCPN mol. \#1 & 810 & 0.43 & 4.2 & - & \cmark & 0 & 0.29 & 22\\
GCPN mol. \#2 & $2.7 \cdot 10^4$ & 0.80 & 3.7 & - & \cmark & 0 & 0.56 & 47\\
\midrule
Tamoxifen & 87 & 0.45 & 2.0 & 1.5* & \cmark & 0 & 0.23 & 16\\
Raloxifene & $7.9 \cdot 10^{6}$ & 0.32 & 2.4 & 0.030* & \cmark & 0 & 0.25 & 59\\
\midrule
\multicolumn{9}{c}{ACAA1}\\
\midrule
LIMO mol. \#1 & \textbf{28} & 0.57 & 5.5 & $\mathbf{4 \cdot 10^{4}}$ & \cmark & 0 & 0.52 & 52\\
LIMO mol. \#2 & 31 & 0.44 & 4.9 & No binding & \cmark & 0 & 0.81 & 45\\
\midrule
GCPN mol. \#1 & 8500 & 0.69 & 4.2 & - & \cmark & 0 & 0.52 & 61\\
GCPN mol. \#2 & 8500 & 0.54 & 4.3 & - & \cmark & 0 & 0.52 & 30\\
\bottomrule
\end{tabular}
\end{sc}
\end{small}
\end{center}
\vskip -0.15in
\end{table*}

\begin{figure*}[h]
\begin{center}
\centerline{\includegraphics[width=0.9\textwidth]{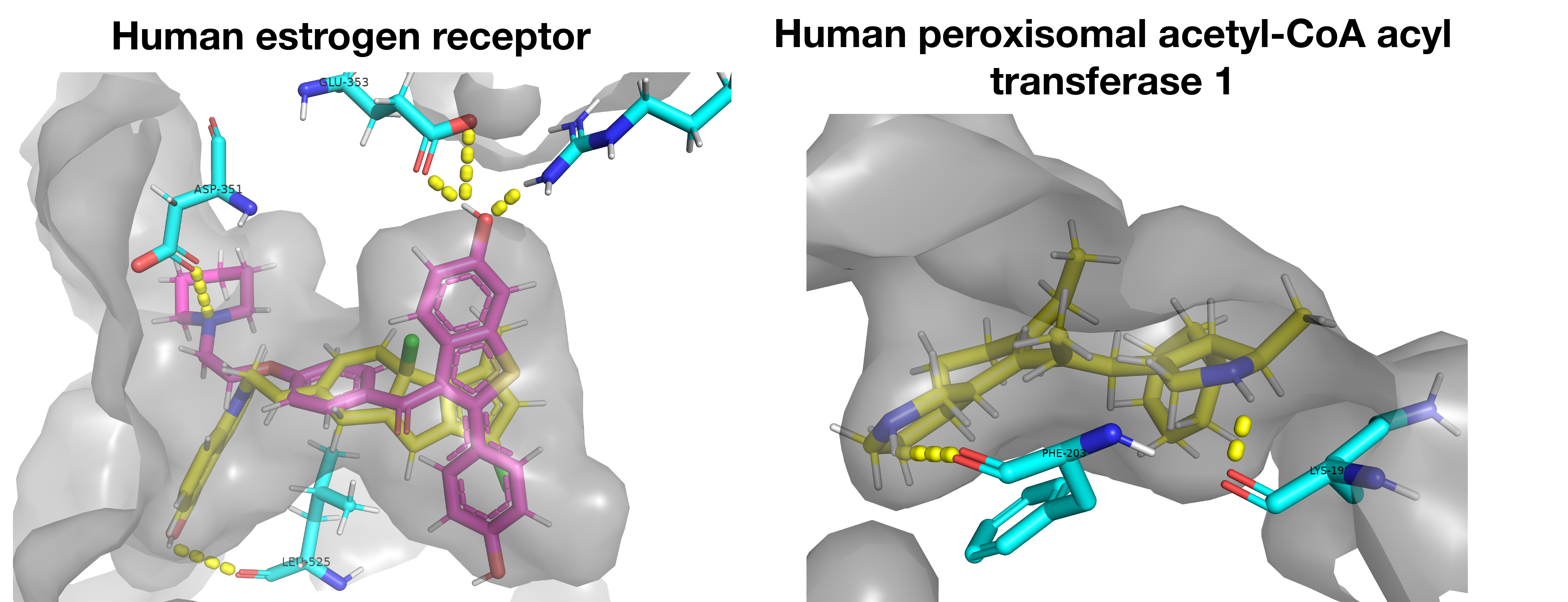}}
\caption{3D visualization of ligands docked against ESR1 and ACAA1. LIMO-generated ligands (one for each protein) are shown in yellow, and raloxifene, a cancer drug that targets ESR1, is shown in pink. The protein pocket is displayed as semi-opaque, and nearby structures of the protein are shown in blue. Docked poses were generated by GLIDE \cite{friesner2004glide} for ESR1 and AutoDock-GPU \cite{santosmartins2021autodockgpu} for ACAA1. Favorable atom-atom interactions between ligand and protein are shown with a dashed yellow line.}
\label{3d-vis}
\end{center}
\vskip -0.2in
\end{figure*}

To generate molecules with high computed binding affinity {\em and} pharmacologic and synthetic desirability, we  simultaneously optimize molecules for computed binding affinity, drug-likeness (QED), and synthesizability (SA) scores. Distributions of properties before and after multi-objective optimization are shown in Appendix \ref{sec:justification}. For each protein target, we generate 100k molecules, then apply the two refinement steps described in Section \ref{sec:refinement}. We selected the two compounds with the highest affinity from this process for each protein target, which are shown in the top row of Figure \ref{mol-examples}. These compounds are more drug-like and synthetically accessible than those generated by single-objective optimization (Figure \ref{mol-examples}, bottom row), but still have high predicted binding affinities (i.e., low $K_D$ values), making them promising drug candidates. We analyze and compare these compounds in the subsequent section.

\paragraph{Compound analysis.}
Table \ref{multi-obj-comparison} shows the binding and drug-likeness metrics of two generated compounds for both ESR1 and ACAA1 (the same as those shown in the top row of Figure \ref{mol-examples}). For ESR1, we compare our compounds to tamoxifen and raloxifene, two modern breast cancer drugs on the market that target this protein. We also compare with compounds generated by GCPN, the second strongest method behind LIMO for single-objective binding affinity maximization, with identical multi-objective weights and the same filtering step as LIMO. For each compound, we report the metrics described in the Appendix \ref{sec:molecule-metrics}. The first three metrics given are ``Optimized properties'' that are explicitly optimized for, while the others are not used in optimization but are still useful for compound evaluation.

LIMO significantly outperforms GCPN, which generates molecules with such low computed affinity (high $K_D$) as to be of relatively low utility in drug discovery, regardless of drug-likeness or synthesizability metrics, because they are unlikely to bind their targets at all.

\paragraph{Visualization and corroboration of binding affinities.}
To confirm that the ligands generated by LIMO are likely to bind their target proteins with high affinity and do not score well due to inaccuracies or shortcuts used in the AutoDock-GPU scoring function, we visualized their docked poses in 3D to look for physically reasonable bound conformations and energetically favorable ligand-protein interactions. The 3D binding poses produced by the docking software for one of the two generated ligands for each protein (Figure \ref{3d-vis}) show that they fit well into the protein binding pocket and promote favorable ligand-protein interactions. 

We furthermore ran detailed, molecular dynamics-based, absolute binding free energy calculations (ABFE, Appendix \ref{sec:abfe-methods}) \cite{gilson1997statsthermobasis, cournia_rigorous_2020} to obtain more reliable estimates of the affinities of LIMO-generated compounds for their targeted proteins than the predictions from docking. As shown in Table \ref{multi-obj-comparison}, LIMO generated an ESR1 ligand with an ABFE-predicted dissociation constant ($K_D$) of $6 \cdot 10^{-5} \text{nM}$, much better than typical $K_D$ values of e.g. 1000 nM obtained from experimental compound screening and better even than the $K_D$ values of tamoxifen and raloxifene, two drugs that bind ESR1 with high affinity. The LIMO compounds  even exceed these drugs on many drug-likeness metrics. Without experimental confirmation, we cannot be sure these molecules bind so well, but the results from these state-of-the-art calculations are encouraging.

\section{Discussion and Conclusions}

We present LIMO, a generative modeling framework for {\em de novo} molecule design. LIMO utilizes a VAE latent space and two neural networks in sequence to reverse-optimize molecular properties, allowing the use of efficient gradient-based optimizers to achieve competitive results on benchmark tasks in significantly less time. The ability to generate six times as many molecules per unit of time relative to competing methods (Table \ref{single-obj-binding}) increases the odds of producing high-quality drug candidates that survive successive rounds of refinement, thereby accelerating drug development as a whole, especially given LIMO's high diversity of compounds (Table \ref{property-targeting}, \ref{random-generation}). On the task of generating molecules with high binding affinity, LIMO outperforms all state-of-the-art baselines.

LIMO promises multiple applications in drug discovery. The ability to quickly generate high-affinity compounds can accelerate target validation with biological probes that can be used to confirm the proposed biological effect of a target. LIMO also has the potential to accelerate lead generation and optimization by jumping directly to drug-like, synthesizable, high affinity compounds, thus bypassing the traditional hit identification, hit-to-lead, and lead optimization steps. While ``unconstrained'' LIMO can quickly generate high-affinity compounds, it has the additional ability to perform substructure-constrained property optimization, which is especially useful during the lead optimization stage where one has an established substructure with a synthetic pathway and wishes to ``decorate'' around it for improved activity or pharmacologic properties.

While LIMO can generate very high affinity compounds as computed by docking software, as is its goal, the utility of compounds only vetted by docking software may be questioned. As shown in Table \ref{multi-obj-comparison}, AutoDock-GPU computed binding affinities do not correlate very well with more accurate ABFE results. This is a well-known result \cite{cournia_rigorous_2020}, but we believe having docking-predicted high affinity compounds is still of relatively high utility in drug discovery, even if some (or most) generated compounds are ``false positives.'' As LIMO can generate hundreds of diverse docking-computed nanomolar range compounds against a target in hours, it is likely that some of those compounds will actually bind a target well. This is a unique advantage of LIMO, as it is able to generate many candidate compounds very quickly, allowing for aggressive filtering downstream. Indeed, we have generated a highly favorable compound ($K_D = 6 \cdot 10^{-14}$ M) as calculated by ABFE, even more favorable than AutoDock-GPU predictions, out of only two generated candidates. The addition of further automated binding affinity confirmation into the LIMO pipeline, e.g. with additional docking software or automated ABFE calculation, is a promising direction for future work. Other future directions include exploring the use of different molecular representation and model architectures in LIMO, the use of better optimizers beyond simple gradient-based methods, and the application of LIMO to more or multiple simultaneous protein targets.
\section*{Acknowledgements}
This work was supported in part by U.S. Department Of Energy, Office of Science, AWS Machine Learning Research Award, and NSF Grant \#2037745. MKG acknowledges funding from National Institute of General Medical Sciences (GM061300). These ﬁndings are solely of the authors and do not necessarily
represent the views of the NIH. MKG has an equity interest in and is a cofounder and scientiﬁc advisor of VeraChem LLC.

\bibliography{main.bib}
\bibliographystyle{icml2022}

\newpage
\appendix
\onecolumn

\section{Experiment description and baselines}

\subsection{Tasks} \label{sec:appendix-tasks}

\textbf{Random generation of molecules}: Generate random molecules by sampling from the latent space of the generative model. As later optimization relies on these generated molecules as starting points, it is important that they be novel, diverse, and unique.

\textbf{QED and penalized logP maximization}: Generate molecules with high penalized logP (p-logP, estimated octanol-water partition coefficient penalized by synthetic accessibility (SA) score and number of cycles with more than six atoms \cite{jin2018jtvae}) and quantitative estimate of drug-likeness (QED, \cite{bickerton2012qed}) scores. These properties are important considerations in drug discovery, and this task shows the ability of a model to optimize salient aspects of a molecule, even if maximization of these properties by themselves is of low utility \cite{zhou2019moldqn}.

\textbf{logP targeting}: Generate molecules with logP within a specified range. In drug discovery, a logP within a given range is often taken as an approximate indicator that a molecule will have favorable pharmacokinetic properties.

\textbf{Similarity-constrained penalized logP maximization}: For each molecule in a set of starting molecules, generate a novel molecule with a high penalized logP (p-logP) score while retaining similarity (as defined by Tanimoto similarity between Morgan fingerprints, \citet{rogers2010morganfingerprint}) to the original molecule. This mimics the drug discovery task of adjustment of an active starting molecule's logP while keeping similarity to the starting molecule to retain biological activity.

\textbf{Substructure-constrained logP extremization}: Generate molecules with either high or low logP scores while keeping a subgraph of a starting molecule fixed. This task mimics the drug discovery goal of optimizing around (``decorating'') an existing substructure to fine-tune activity or adjust pharmacologic properties. This is common in the lead optimization stage of drug development, where a synthetic pathway to reach an exact substructure with proven activity is established, but molecular groups around this substructure are more malleable and not yet established. This task is not captured in the similarity-constrained optimization task above, which uses more general whole-molecule similarity metrics.

\textbf{Single-objective binding affinity maximization}: Generate molecules with high computed binding affinity for two protein targets as determined by docking software. Reaching high binding affinities is the primary goal of early drug discovery, and its optimization using a physics-based affinity estimator is a relatively novel task in the ML-based molecule generation literature. Previous attempts to optimize affinity have relied on knowledge of existing binders \cite{zhavoronkov2019gentrl, jeon2020autonomous, luo20213d}, which lacks the generalizability of physics-based estimators to targets without known binders.

\textbf{Multi-objective binding affinity maximization}: Generate molecules with favorable computed binding affinity, QED, and SA scores. This task has high utility in drug discovery, as it addresses targeting, pharmacokinetic properties, and ease of synthesis. Development of molecules satisfying all these considerations is challenging, and to the best of our knowledge, is a novel task in the ML-based molecule generation literature.

\subsection{Molecule metrics} \label{sec:molecule-metrics}
We report the following metrics for our multi-objective optimized molecules, all of which are given by ADMETLab 2.0 \cite{xiong2021admetlab20} except binding affinities:

\begin{itemize}
    \item \textbf{$\boldsymbol{K_D}$ (AutoDock-GPU)}: Dissociation constant $K_D$ in nanomolar, as calculated by AutoDock-GPU. Lower $K_D$ is associated with better binding (i.e. higher affinity) \cite{santosmartins2021autodockgpu}
    \item \textbf{$\boldsymbol{K_D}$ (ABFE)}: Dissociation constant $K_D$ in nanomolar, as calculated by absolute binding free energy (ABFE) calculations, which are generally more accurate than AutoDock-GPU scores \cite{cournia_rigorous_2020}
    \item \textbf{QED}: Quantitative estimate of drug-likeness score, higher is better \cite{bickerton2012qed}
    \item \textbf{SA}: Synthetic accessibility score, lower is better \cite{ertl2009sa}
    \item \textbf{Lipinksi}: Lipinski's rule of 5 is a commonly used rule of thumb for drug-likeness \cite{lipinski2001rule}. Compounds that pass all or all but one of four components are considered more likely to be suitable as drugs.
    \item \textbf{PAINS}: Number of PAINS alerts. These alerts detect compounds likely to have non-specific activity against a wide array of biological targets, making them undesirable as drugs. Lower is better \cite{baell2010pains}
    \item \textbf{$\text{Fsp}^3$}: The fraction of sp3 hybridized carbons, which is thought to correlate with favorable drug properties. Higher is better \cite{wei2020fsp3}
    \item \textbf{MCE-18}: A measure of molecular novelty based on complexity measures. Higher is better \cite{ivanenkov2019mce18}
\end{itemize}

\subsection{Fine-tuning algorithm} \label{sec:appendix-finetuning}

\begin{algorithm}[h]
\caption{Molecule fine-tuning algorithm.}
\begin{algorithmic}
    \REQUIRE The starting molecule $\mathcal{M}$ to be optimized and a function $\pi(m)$ that calculates a property for molecule $m$
    \STATE $\mathcal{R} \gets \{\text{N}, \text{O}, \text{Cl}, \text{F}\}$
    \STATE $\text{bestProperty} \gets \pi(\mathcal{M})$
    \WHILE{bestProperty is improving}
    \STATE $\text{bestMolecule} = \mathcal{M}$
    \FORALL{$\text{carbon atoms in } \mathcal{M} \text{ not adjacent to any atoms } \in \mathcal{R}$}
    \FORALL{potential replacement atoms $\in \mathcal{R}$}
    \STATE $m \gets \mathcal{M} \text{ with the carbon atom replaced}$
    \IF{$m$ is valid and $\pi(m)$ is better than $\pi(\text{bestMolecule})$}
    \STATE $\text{bestMolecule} \gets m$
    \ENDIF
    \ENDFOR
    \ENDFOR
    \STATE $\mathcal{M} \gets \text{bestMolecule}$
    \ENDWHILE
\end{algorithmic}
\end{algorithm}

\subsection{Baselines} \label{sec:appendix-baselines}

We compare with the following baselines:

\begin{itemize}
    \item JT-VAE \cite{jin2018jtvae}: a VAE-based generative model that first generates a scaffold junction tree and then assembles nodes in the tree into a molecular graph.
    \item GCPN \cite{you2018gcpn}: an RL agent that successively constructs a molecule by optimizing a reward composed of molecular property objectives and adversarial loss. For running baselines, we use code from \url{https://github.com/bowenliu16/rl_graph_generation}.
    \item MolDQN \cite{zhou2019moldqn}: an RL framework that uses chemical domain knowledge and double Q-learning. For running baselines, we use code from \url{https://github.com/aksub99/MolDQN-pytorch}.
    \item MARS \cite{xie2021mars}: a sampling method based on Markov chain Monte Carlo that uses an adaptive fragment-editing proposal distribution based on GNN. 
    \item GraphDF \cite{luo2021graphdf}: a normalizing flow model for graph generation that uses a discrete latent variable model, fine-tuned with RL. For running baselines, we use code from \url{https://github.com/divelab/DIG}.
\end{itemize}

To generate results from baselines, we ran each method until little improvement was observed. For methods without an explicit generation process (i.e. GCPN, MolDQN, and MARS), we took the highest property scores from all molecules generated. For methods with an explicit generation process (GraphDF), we trained until little improvement was observed and then sampled the same number of molecules as was sampled from LIMO. All times reported include the total time from each method, including training, property calculation times, and generation times if applicable.

To run MolDQN for the property targeting task, which requires obtaining an optimized set of molecules, we used the last molecule of the most recent 1,000 training episodes to build a set on which success and diversity were calculated.

\subsection{Experimental details} \label{sec:experimental-details}

For the VAE, we use a 64-dimensional embedding layer that feeds into four batch-normalized 1,000-dimensional (2,000 for first layer) linear layers with ReLU activation. This generates a Gaussian output for the 1024-dimensional latent space that can be sampled from. For the decoder, we also use four batch-normalized linear layers with ReLU activation, with the same dimensions. Softmax is used over all possible symbols at each symbol location in the output layer, and the VAE is trained with evidence lower bound (ELBO), with the negative log-likelihood reconstruction term multiplied by 0.9 and the KL-divergence term multiplied by 0.1. The VAE is trained over 18 epochs with a learning rate of 0.0001 using the Adam optimizer.

For the property predictor, we use three 1,000-dimensional linear layers with ReLU activation. Layer width, number of layers, and activation function were determined after hyperparameter optimization for predicting penalized logP, and these hyperparameters were then used for all other property prediction tasks. Similarly, we did not tune baseline methods for specific tasks, and used only the default hyperparameters tuned on a single task. For each property to predict, we use PyTorch Lightning to choose the optimal learning rate with the Adam optimizer to train the predictor over 5 epochs, then perform backward optimization with a learning rate of 0.1 for 10 epochs.

All experiments, including baselines, were run on two GTX 1080 Ti GPUs, one for running PyTorch code and the other for running AutoDock-GPU, and 4 CPU cores with 32 GB memory.

\subsection{Autodock-GPU} \label{sec:autodock-methods}

We use AutoDock-GPU, a GPU accelerated version of AutoDock4 with an additional AdaDelta local search method, to calculate binding affinities for LIMO. It is fast enough for our purposes while still generating reasonably accurate results \cite{santosmartins2021autodockgpu}. 

To generate docking scores from a SMILES string produced by LIMO, we perform the following steps:

\begin{enumerate}
    \item Generate grid files for docking using AutoGrid4. For human estrogen receptor, we set the bounding box for docking to include the well-known ligand binding site. For human peroxisomal acetyl-CoA acyl transferase 1, a novel target, we use fpocket \cite{leguilloux2009fpocket} to predict the binding pocket and set the docking bounding box around it.
    \item For each SMILES to evaluate, we convert it to a 3D .pdbqt file using obabel 2.4.0 \cite{oboyle2011obabel}. We set pH=7.4 to assign hydrogens and set Gasteiger partial charges.
    \item We run AutoDock-GPU \cite{santosmartins2021autodockgpu} with default parameters on the .pdbqt files, in batch mode if applicable.
    \item With the generated .dig files from AutoDock-GPU, we extract the top binding energy number in the results table.
\end{enumerate}

\subsection{Absolute binding free energy} \label{sec:abfe-methods}
To corroborate our AutoDock-GPU predicted binding affinities, we conducted absolute binding free energy (ABFE) calculations on our most promising ligands. ABFE calculations estimate the binding free energy $\Delta G_{bind}$, i.e., the difference between the free energy of a molecule's bound and unbound states, by computing the reversible work of moving a molecule from water into the binding site of the targeted protein. The dissociation constant is then obtained as $K_{D}(ABFE) = e^{-\Delta G_{bind}/RT}$, where $R$ is the gas constant and $T$ is absolute temperature \cite{gilson2007abfe}. The free energy calculation is done with detailed molecular dynamics simulations of the protein and the molecule dissolved in thousands of water molecules.  This method is more detailed and computationally expensive, and typically more accurate, than docking \cite{cournia_rigorous_2020}. Here, the 5 best-scoring poses from AutoDock-GPU were sent to the software BAT.py \cite{heinzelmann_automation_2021} to compute the binding free energy, $\Delta G_i$, for each pose $i$. The overall binding free energy accounting for all 5 poses was then obtained as $\Delta G_{bind}=-RT \ln \sum_i e^{-\Delta G_i/RT}$ \cite{gilson2007abfe}. Note that the pose with the most favorable (negative) $\Delta G_i$ contributes the most to the overall binding free energy, and this is also the most stable and hence most probable binding pose of the ligand. We thus analyzed the protein-ligand interactions for this most stable pose. For each ligand, we use the mean free energy of two independent ABFE runs from calculations initiated with different random seeds.

\section{Additional experiments}
\subsection{Random generation of molecules} \label{sec:appendix-random-generation}

\begin{table}[ht]
    \caption{Random generation of molecules trained on the MOSES dataset and calculated with the MOSES platform \cite{polykovskiy2020moses}. \% valid: percent of molecules that are chemically valid. U@1K: percent of 1,000 generated molecules that are unique. U@10K: percent of 10,000 generated molecules that are unique. Diversity: one minus average pairwise similarity between molecules. \% novel: percent of valid generated molecules not present in training set. JT-VAE results taken from \cite{polykovskiy2020moses}.}
    \label{random-generation}
    \tabcolsep=0.10cm
    \vskip 0.1in
    \begin{center}
    \begin{small}
    \begin{sc}
    \begin{tabular}{lccccc}
    \toprule
    Method & \% valid & \% U@1k & \% U@10k & Div. & \% Nov.\\
    \midrule
    JT-VAE & \textbf{100} & \textbf{100} & \textbf{99.96} & 0.855 & 91.43\\
    GraphDF & \textbf{100} & \textbf{100} & 99.72 & 0.887 & \textbf{100}\\
    \midrule
    LIMO & \textbf{100} & 99.8 & 97.56 & \textbf{0.907} & \textbf{100}\\
    \bottomrule
    \end{tabular}
    \end{sc}
    \end{small}
    \end{center}
    \vskip -0.1in
    \end{table}

Results from the random generation of 30,000 molecules are summarized in Table \ref{random-generation}. LIMO achieves the highest diversity score among compared methods, an important metric when using the latent space as a basis for the optimization of molecules on a wide range of properties. This diversity provides the foundation for LIMO's ability to generate a diverse set of molecules with desirable properties.

\subsection{Justification of multi-objective optimization} \label{sec:justification}

\begin{figure*}[h]
\begin{center}
\centerline{\includegraphics[width=1.0\textwidth]{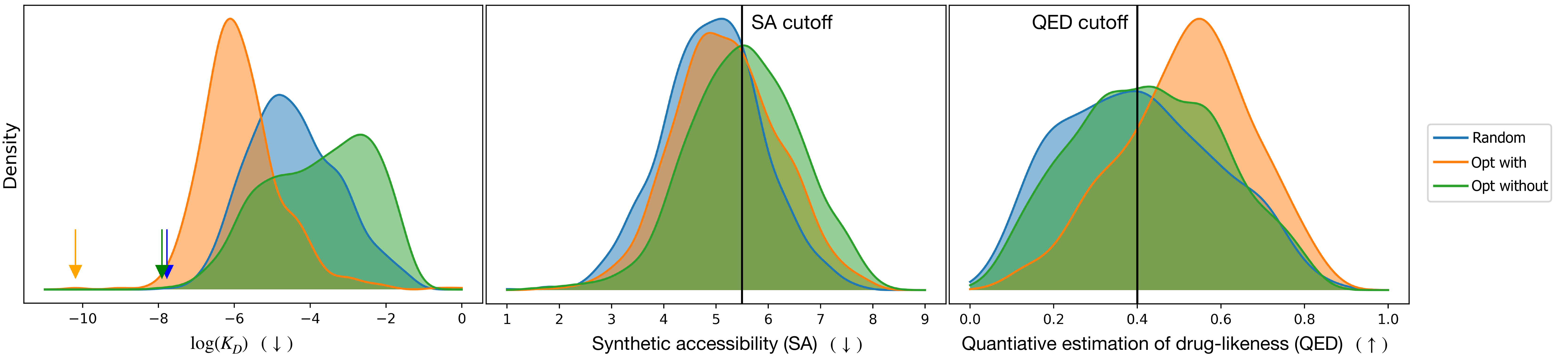}}
\caption{Distribution of molecular properties of randomly generated molecules (Random), after performing optimization on all three properties (Opt with), and after performing optimization on the two other properties, leaving out the one on the x-axis (Opt without). For QED and SA, cutoff values are shown for the minimum and maximum (respectively) scores that we consider sufficient for further optimization. For the $K_D$ distributions, arrows mark the minimum value of each. On the x-axis, ($\downarrow$) indicates that a low value is desired, and ($\uparrow$) indicates that a high value is desired.}
\label{distribution-change}
\end{center}
\vskip -0.2in
\end{figure*}

Figure \ref{distribution-change} shows distributions of properties from randomly sampled molecules, molecules optimized on all three objectives (computed binding affinity against ESR1, QED, and SA), and optimized molecules leaving out one objective. We also show our QED and SA cutoff values used in the filtering step defined in Section \ref{sec:refinement}. As shown, inclusion in the objective function pushes each property in the direction of improvement, or, in the case of SA, prevents it from decreasing more than it would have if it had not been included. Therefore, multi-objective optimization is successful in generating more molecules with potentially high binding affinity within the defined cutoff ranges, so is advantageous over single-objective optimization.

\end{document}